\documentclass[conference]{IEEEtran}
\IEEEoverridecommandlockouts
\ifCLASSINFOpdf
  \usepackage[pdftex]{graphicx}
  \graphicspath{{../pdf/}{/jpeg/}}
  \DeclareGraphicsExtensions{.pdf,.jpg,.png}
\else
  \usepackage[dvips]{graphicx}
\graphicspath{{../eps/}}
  \DeclareGraphicsExtensions{.eps}
\fi
\usepackage{algorithmic}
\usepackage{algorithmic}
\usepackage{algorithm}

\begin{document}
%
\title{DRE-Bot: A Hierarchical First Person Shooter Bot Using Multiple Sarsa($\lambda$) Reinforcement Learners}

\author{\IEEEauthorblockN{Frank G. Glavin}
\IEEEauthorblockA{College of Engineering \& Informatics,\\
National University of Ireland, Galway,\\
Ireland.\\
Email: frank.glavin@nuigalway.ie}
\and
\IEEEauthorblockN{Michael G. Madden}
\IEEEauthorblockA{College of Engineering \& Informatics,\\
National University of Ireland, Galway,\\
Ireland.\\
Email: michael.madden@nuigalway.ie}
}


%


\maketitle

\begin{abstract}
This paper describes an architecture for controlling non-player characters (NPC) in the First Person Shooter (FPS) game Unreal Tournament 2004. Specifically, the DRE-Bot architecture is made up of three reinforcement learners, Danger, Replenish and Explore, which use the tabular Sarsa($\lambda$) algorithm. This algorithm enables the NPC to learn through trial and error building up experience over time in an approach inspired by human learning. Experimentation is carried to measure the performance of DRE-Bot when competing against fixed strategy bots that ship with the game. The discount parameter, $\gamma$, and the trace parameter, $\lambda$, are also varied to see if their values have an effect on the performance.
\end{abstract}

\begin{IEEEkeywords}
 Reinforcement Learning, First Person Shooter
\end{IEEEkeywords}

%
\IEEEpeerreviewmaketitle

\section{Introduction}
\subsection{Reinforcement Learning}
Reinforcement learning (RL) \cite{RLIntro} involves a decision-making learner, often referred to as an agent, which carries out actions in an environment in order to achieve an explicit goal or goals. The list of all possible actions that can be taken is called the \emph{action space} and the list of all states is known as the \emph{state space}. State-action pairs are stored by the \emph{policy} of the learner. These represent how useful it is to carry out a specific action in a given state. The policy, which is informed by a numerical reward received at each time step, adapts over time with a view to maximise the long term reward. The most common types of policy in reinforcement learning are generalisation and tabular approaches. With generalisation, a function approximator is used to generalise a mapping of states to actions. The tabular approach stores numerical representations of all state-action pairs in a lookup table. In this research, we use the tabular Sarsa($\lambda$) algorithm which is described in Section III.

\subsection{First Person Shooter Games}
FPS games are a genre of computer game in which players are immersed in a competitive, three dimensional environment. Players are controlled from a first person perspective and can traverse through the world, interacting with both the environment and other players (team mates and opponents). Many different game types exist with FPSs. In this paper, we are only concerned with developing a bot that is competent in playing the Death Match or ``Free-for-all" game type in which the sole objective for each player taking part is to eliminate the other players in the world. Players can pick up items such as guns, ammunition and health packages and engage in combat with each other until either a score or time limit is reached. The skills acquired in this mode are, of course, prerequisites for the other game types.

\subsection{Problem Summary}
Graphics in modern computer games are closer than ever to being photo-realistic. In recent times, there has been an increased emphasis on improved artificial intelligence (AI) with game developers realising its importance for making games more adaptable and enjoyable. Traditional approaches to game-AI included hard-coding, scripting techniques such as Finite State Machines and Fuzzy logic. While these techniques are tried and tested with varying success, they can often lead to predictable game play that can be exploited by knowledgeable players. Controlling an FPS bot in a complex 3D environment is certainly a difficult task, even for a human. Constant real time decision making is required for a variety of different tasks such as path-finding, combat, retreating, completing game objectives etc. The long-term objective of this work is to develop bots that can learn their own strategy and continually adapt over time, as opposed to being given strict rules, to test whether this will lead to a more challenging and natural opponent for players of all skill levels.

\section{Related Work}
BotPrize \cite{TurBot09} is a competition that was set up for testing the humanness of computer controlled bots in FPS games. The initial purpose of the competition was to see if computer controlled bots could fool expert judges into believing that they were human players in Unreal Tournament 2004. This competition essentially acts as a Turing Test \cite{Turing} for bots. The overall goal of the competition, which was first run in 2008, was to fool the expert judges into believing that a bot is human at least 50 percent of the time. The format of the competition has since changed \cite{TurBot10} and now all players, including bots, act as judges in the game by using a modified gun to carry out their judgement of the other players.\\
\indent McPartland and Gallagher \cite{RLFPS} applied the tabular Sarsa($\lambda$) \cite{RLIntro} algorithm to a purpose-built first person shooter game. The algorithm was used to learn the controllers of navigation, item collection and combat individually. The authors report that they developed a purpose built game as opposed to using a commercial game in order to reduce the processor overhead and increase the throughput of experiments. The results showed that reinforcement learning could be successfully applied to a simplified purpose-built FPS game. \\
\indent Smith et al. \cite{Retal} developed an algorithm called RETALIATE (REinforced TActic Learning In Agent-Team Environments) for Unreal Tournament. The authors used the Q-learning \cite{Qlearn} algorithm for learning winning policies in the Domination game type. The work that was carried out was concerned with co-ordinating the team behaviour, as opposed to learning behaviours of the individual players. They carried out experiments against three different teams with varying strategies. The results showed that the algorithm adapted well to the changing environments. This algorithm was later enhanced using Case Based Reasoning by Auslander et al. \cite{RLCBR} with their agent called CBRetaliate. \\
\indent Di Wang et al. \cite{RLFPS2} proposed the use of FALCON \cite{Falcon} for developing a computer-controlled agent in Unreal Tournament 2004. The authors built two FALCON networks, one for weapon selection and one for behaviour selection. The bot learned by using cognitive nodes which could be translated into rules by associating a state and a particular action with an estimated reward. The bots created these rules and learned how to play the game in real time.

\section{Game and Development Software}
\subsection{Unreal Tournament 2004}
Unreal Tournament 2004 is a highly customisable, commercial FPS game that was developed by Epic Games and Digital Extremes. Players can choose from a large variety of avatars and maps competing against other human players, computer controlled bots or a mixture of both. The game uses the Unreal Engine which has an open scripting language called UnrealScript. This scripting language can be used to carry out high level programming of the game.

\subsection{Pogamut 3}
Pogamut 3 is an open-source Java middleware toolkit\footnote{http://pogamut.cuni.cz/main/tiki-index.php} which can be used for creating virtual agents in the 3D environment of Unreal Tournament 2004. It makes use of UnrealScript for developing external control mechanisms for the game. The toolkit enables developers to spawn and control bots and has simplified the process of carrying out in-game actions, such as path-finding, so that the development emphasis can be put on the intelligence of the bot.

\section{DRE-Bot Architecture}

\subsection{Overview}
We have developed three high level modes for this architecture. These are \emph{Danger}, \emph{Replenish} and \emph{Explore} mode. \emph{Danger} mode is activated when the bot can see an opponent or is being damaged. \emph{Replenish} mode is activated when ammunition and/or health is below a low (40\%) or critical (20\%) level. \emph{Explore} mode is activated when the bot is not in \emph{Danger} or \emph{Replenish} mode. Each of these modes consists of one independent learner which has its own states, actions and rewards. 
\begin{figure}[h]
  \centering
  \includegraphics[width=2in]{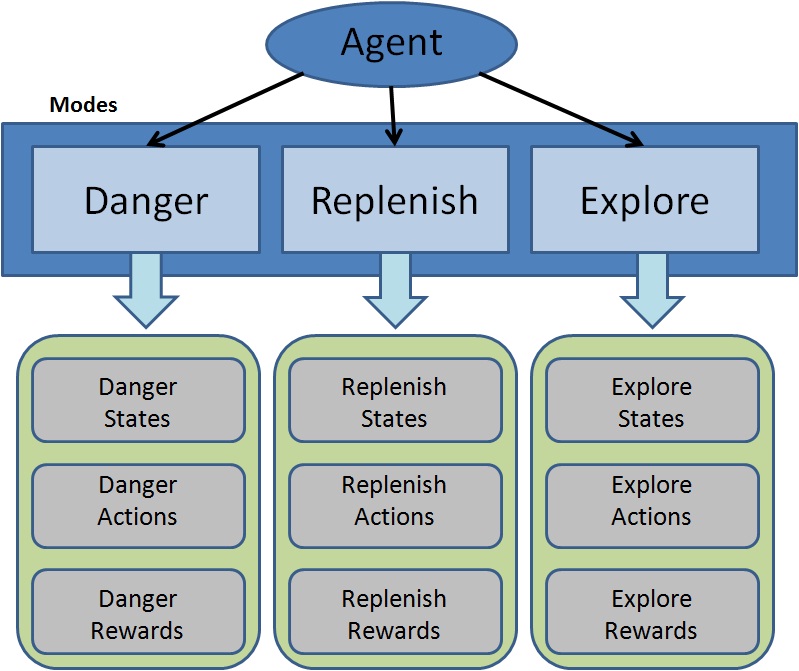}
  \caption{DRE-Bot Architecture.}
\end{figure}
Each of the learners in this implementation uses the Sarsa ($\lambda$) algorithm, which will be described in the next section, and has its own state-action lookup table. The key motivation for this architecture is to decompose the problem space, so as to improve learning. A benefit of this architecture is that different learning algorithms or varying parameters could be used for the different modes if it was deemed to be advantageous to the learning process. However, for this current implementation, the same algorithm and parameters are used by all three modes.

\subsection{Tabular Sarsa($\lambda$) Algorithm}
The Tabular Sarsa($\lambda$) algorithm learns directly from raw experience without any model of the environment's dynamics. Being an on-policy method, the algorithm continually estimates state-action values (Q-values) for a specific behaviour policy while, at the same time, changing toward greediness with respect to the Q-values. It uses eligibility traces to speed up learning by allowing past actions to benefit from the current reward. The use of eligibility traces also enables the algorithm to learn sequences of actions which can be beneficial when learning useful policies in first person shooter games. The pseudo-code for the algorithm is written in Fig. 2.
\begin{figure}
\caption{Pseudo-code for the Sarsa($\lambda$) algorithm.}
\begin{algorithmic}
\STATE  Initialise \emph{Q}(\emph{s}, \emph{a}) = 0, \emph{e}(\emph{s}, \emph{a}) = 0 for all \emph{s}, \emph{a}
\REPEAT 
\STATE Initialise \emph{s}
\STATE Choose \emph{a} from \emph{s} using policy derived from \emph{Q} 
\REPEAT
\STATE Take action \emph{a}, observe \emph{r}, \emph{s'}
\STATE Choose \emph{a'} from \emph{s'} using policy derived from \emph{Q} 

\STATE $\delta$ $\Leftarrow$ \emph{r} + $\gamma$\emph{Q}(\emph{s'}, \emph{a'}) - \emph{Q}(\emph{s}, \emph{a})
\STATE \emph{e}(\emph{s}, \emph{a}) $\Leftarrow$  1
\STATE For all \emph{s}, \emph{a}:
\STATE\hspace{\algorithmicindent} \emph{Q}(\emph{s}, \emph{a})  $\Leftarrow$ \emph{Q}(\emph{s}, \emph{a}) + $\alpha$$\delta$\emph{e}(\emph{s}, \emph{a})
\STATE\hspace{\algorithmicindent} \emph{e}(\emph{s}, \emph{a}) $\Leftarrow$ $\gamma$$\lambda$\emph{e}(\emph{s}, \emph{a})
\STATE \emph{s} $\Leftarrow$ \emph{s'}; \emph{a} $\Leftarrow$ \emph{a'}
\UNTIL{(steps of single episode have finished)}
\UNTIL{(all episodes have finished)} 
\end{algorithmic}
\end{figure}
Firstly, the Q-values and eligibility traces for all states and actions are initialised. Then, for every step of each episode, \emph{s’} is set to the current state and an available action \emph{a’} is selected. The action is taken and then a reward, \emph{r}, is received. The temporal difference (TD) error, $\delta$, is then calculated using the reward, discount parameter, $\gamma$, and the current and next state-action pairs. The current eligibility trace is then assigned a value of 1 to mark it as being eligible for learning. Next, the Q-values and eligibility traces for all states and actions are updated as follows. Each Q-value is updated as the old Q-value plus the eligibility trace variable multiplied by the learning rate, $\alpha$, and the TD error $\delta$. Each eligibility trace variable is then updated as the old value multiplied by the discount parameter ($\gamma$) and the eligibility trace ($\lambda$). Once this has completed, the current state s is set to the next action \emph{s'} and the current action, \emph{a}, is set to the next action \emph{a'}.

\subsection{States, Actions and Rewards}
\subsubsection{States}
The states are made up of a series of checks from the environment. Checks such as ``Is an opponent visible?'', ``Am I being hit?'', ``Do I have low health?'' are combined together to make up the individual states. All of these checks correspond to the bots perception of the world around it. Every new state that is added increases the complexity of the learner. For this reason, it was important to design high level states that capture the most important information from the game. Too few states could lead to consistently poor performance where as too many could drastically increase the amount of time it takes to learn an effective policy. The following are tables of the checks that are carried out to represent the states in each of the modes. \\
\indent Firstly, in Table I, we can see the different checks and values which are used to represent the \emph{Danger} state. Being hit, bumping and hearing noise can each have the value of either True or False. The distance check measures how far an opponent is from the bot. These values are \emph{short}, \emph{medium}, \emph{far} or \emph{no} which represents that no opponent is visible at the moment. There are 32 different \emph{Danger} states in total. 

\begin{table}[h]
\caption{DRE-Bot Danger states}
\begin{center}
\begin{tabular}{|c|c|}
   \hline
   \bf{Check}  & \bf{Values} \\
   \hline
  \emph{BeingHit} & True / False \\
\hline
   \emph{Bumping} & True / False \\
\hline
  \emph{HearingNoise} & True / False \\
\hline
   \emph{Distance} & short / medium / far / no\\
   \hline
 \end{tabular}

\end{center}
\end{table}

\indent The checks and values for the \emph{Replenish} states are shown in Table II. These include checks for seeing an enemy, seeing a pickup and hearing a pickup. Poor levels of health and ammunition are also taken into account. There are two levels for health: \emph{low} (LH) and \emph{critical} (CH). There are also two levels for ammunition: \emph{low} (LA) and \emph{critical} (CA). Low corresponds to having 40\% remaining where as critical corresponds to having 20\% remaining. There are 64 possible states when in \emph{Replenish} mode. \\

\begin{table}[h]
\caption{DRE-Bot Replenish states}
\begin{center}
\begin{tabular}{|c|c|}
   \hline
   \bf{Check} & \bf{Values} \\
   \hline
   \emph{SeeEnemy} & True / False \\
\hline
   \emph{SeePickup} & True / False \\
\hline
   \emph{HearPickup} & True / False \\
\hline
   \emph{Levels} & LA/LH/LA\&LH/CA/CH/\\
   & CA\&CH/CA\&LH/LA\&CH \\
   \hline
 \end{tabular}

\end{center}
\end{table}

\indent The final set of checks and values are for the \emph{Explore} states. These are shown in Table III. There are only two checks carried out with 6 possible states altogether. The movement check can return: \emph{walking}, \emph{running} or \emph{stopped}. Whether or not the bot is crouched is also checked.

\begin{table}[h]
\caption{DRE-Bot Explore states}
\begin{center}
\begin{tabular}{|c|c|}
   \hline
   \bf{Check}  & \bf{Values} \\
   \hline
   \emph{Movement} & Walk / Run / Stop \\
 \hline
   \emph{Crouched} & True / False \\
   \hline
 \end{tabular}

\end{center}
\end{table}

\subsubsection{Actions}
The actions that are available to the bot are not primitive but they are also not overly complex in that each action carries out just one activity. These include activities such as continuous movement, jumping, shooting and changing weapon. The amount of actions and their corresponding complexity will determine the eventual proficiency of the bot. We will now take a look at the actions that are available for the three different modes. These high level actions were chosen based on human knowledge of the game and were improved over time after running initial experiments.\\
\indent In \emph{Danger} mode, the bot can shoot its primary weapon, shoot its secondary weapon, go to the location of the last opponent seen (if any), stop all movement, dodge in a random direction, jump with a random amount of elevation, turn and face a visible player or turn randomly, and change weapon.

\begin{table}[h]
\caption{DRE-Bot Danger Actions}
\begin{center}
\begin{tabular}{|c|p{4cm}|}
   \hline
\bf{Action} & \bf{The bot will:} \\
 
  \hline
\emph{ShootPrimary} & shoot player in primary mode. \\
\hline
\emph{ShootSecondary} & shoot player in secondary mode. \\
\hline
\emph{LastSeenOpponent} & go to the last seen opponent. \\
\hline
\emph{StopMovement} & stop all movement completely. \\
\hline
\emph{Dodge} & perform dodging maneuver. \\
\hline
\emph{Jump} & perform random jump. \\
\hline
\emph{FacePlayerOrTurn} & face opponent or turn randomly. \\
\hline
\emph{ChangeWeapon} & change to a different weapon.\\
\hline
 \end{tabular}

\end{center}
\end{table}

There are also some hard-coded rules about taking actions such as not being able to shoot a gun without an opponent being visible. The \emph{Danger} actions are designed to deal with scenarios in which the bot is being damaged and in danger of being killed. \\
\indent The \emph{Replenish} actions are listed below. These include shooting the primary weapon, shooting the secondary weapon, continuous movement, going to a visible pickup, recording the location of visible item, going to an item from those that are stored and turning away from a visible opponent to escape.  

\begin{table}[h]
\caption{DRE-Bot Replenish Actions}
\begin{center}
\begin{tabular}{|c|p{4cm}|}
\hline
\bf{Action} & \bf{The bot will:} \\   
\hline
\emph{ShootPrimary} & shoot player in primary mode. \\
\hline
\emph{ShootSecondary} & shoot player in secondary mode. \\
\hline
\emph{Move} & move continuously straight ahead. \\
\hline
\emph{GoToPickup} & go to pickup, if visible. \\
\hline
\emph{RecordItem} & record location of visible pickup. \\
\hline
\emph{GoToKnownItem} & go to recorded pickup location. \\
\hline
\emph{EscapeOpponent} & turn from opponent and run. \\
\hline
 \end{tabular}

\end{center}
\end{table}

These actions are designed to deal with scenarios in which the bot has low health or ammunition and needs to replenish their supplies. Shooting actions were also included so that the bot could defend itself in such situations. \\
\indent The final table of actions, below, lists the actions that are available to the bot while in Explore mode. These include constant motion around the map while running, constant motion around the map while walking, turning left, turning right, stopping all movement, crouching and un-crouching. This mode is designed for when the bot is not in any danger and has good health and ammunition. 

\begin{table}[h]
\caption{DRE-Bot Explore Actions}
\begin{center}
\begin{tabular}{|c|p{4cm}|}
\hline
\bf{Action} & \bf{The bot will:} \\
   \hline
\emph{RunAround} & move continuously while running. \\
\hline
\emph{WalkAround} & move continuously while walking. \\
\hline
\emph{TurnLeft} & turn left by a random amount. \\
\hline
\emph{TurnRight} & turn right by a random amount. \\
\hline
\emph{StopMovement} & stop all movement completely. \\
\hline
\emph{Crouch} & go into crouched position. \\
\hline
 \end{tabular}

\end{center}
\end{table}

\subsubsection{Rewards}
The reward is acquired through an accumulation of reward checks. These are listed below with their corresponding values. If a reward check returns True then the value for that reward is added to the total. The reward is then returned as a single number which represents the reward received for that time step.

\begin{table}[h]
\caption{DRE-Bot Reward Signal}
\begin{center}
\begin{tabular}{|c|c|}
   \hline
\emph{isHealthy} & + 0.0001 \\
\hline
\emph{isNotHealthy} & - 0.0001 \\
\hline
\emph{isNotColliding} & + 0.00001 \\
\hline
\emph{isColliding} & - 0.00001 \\
\hline
\emph{isMoving} & + 0.00001 \\
\hline
\emph{isNotMoving} & - 0.00001 \\
\hline
\emph{seeOpposingPlayer} & + 0.0001 \\
\hline
\emph{isCausingDamage} & + 0.1\\
\hline
\emph{isBeingDamaged} & - 0.1 \\
\hline
\emph{killedOpponent} & + 1 \\
\hline
\emph{killedByOpponent} & - 1 \\
\hline
\emph{pickedUpItem} & + 0.1 \\
\hline
\emph{gainedAdrenaline} & + 0.2 \\
\hline
 \end{tabular}

\end{center}
\end{table}

The rewards checks include: being healthy or not, colliding or not, moving or not, seeing an opposing player, causing or receiving damage, killing or being killed and picking up items or gaining adrenaline.  Adrenaline can be gained by either picking up pills on the map or completing tasks such as a killing spree (multiple kills) or ending an opponent’s killing spree etc. Gaining adrenaline is given a substantial reward as it is indicative of good play in the game.

\section{Experimentation}
\subsection{Details}
These experiments involve connecting the DRE-Bot to a death match server against two fixed-strategy bots from the game. The games were played on a small map designed for two to three players and the games ended when the DRE-Bot had died 200 times. Both the discount parameter, $\gamma$, and the eligibility trace parameter, $\lambda$ were varied to note what effect they had on the overall performance of the Sarsa($\lambda$) algorithm. The learning rate, $\alpha$, was fixed at 0.2 for all games and an $\epsilon$-greedy strategy with $\epsilon$ set 0.2 was used. Setting $\epsilon$ to 0.2 means that random actions will be chosen 2 out every 10 times.
\subsection{Results and Analysis}
Table VIII below shows the total reward for both runs with the varied discount ($\gamma$) and trace ($\lambda$) parameters. Each run consists of 16 different games. The least amount of reward received in the first run was 164.35 ($\gamma$: 0.6, $\lambda$: 0.6) and the most was 540.99 ($\gamma$: 0.0, $\lambda$: 0.9). The average reward received of all of the 16 games was 382.05.
\begin{table}[h]
\caption{Total Reward Received}
\begin{center}
\begin{tabular}{|c|c|c|c|c|c|}
\hline
 \bf{Run} & & \bf{$\lambda$: 0.0} & \bf{$\lambda$: 0.3} & \bf{$\lambda$: 0.6} & \bf{$\lambda$: 0.9} \\   
\hline
1&\bf{$\gamma$: 0.0}  &448.83  &488.39 &419.72 &540.99 \\
\hline
2&\bf{$\gamma$: 0.0}  &188.57 &153.80 &603.42 &434.43 \\
\hline
\hline
1&\bf{$\gamma$: 0.3} &440.82  &323.68 &195.56 &366.73 \\
\hline
2&\bf{$\gamma$: 0.3} &419.70 &211.76 &470.61 &437.84 \\
\hline
\hline
1&\bf{$\gamma$: 0.6} &485.05  &319.34 &164.35 &325.93 \\
\hline
2&\bf{$\gamma$: 0.6} &362.63 &514.00 &466.32 &397.27 \\
\hline
\hline
1&\bf{$\gamma$: 0.9} &361.74  &476.38 &407.21 &348.08 \\
\hline
2&\bf{$\gamma$: 0.9} &376.39 &469.17 &221.07 &362.64 \\
\hline
 \end{tabular}
\end{center}
\end{table}
A similar figure could be seen for the second run with the average reward from the 16 games being 380.60. However, this time the values ranged from 153.80 ($\gamma$: 0.0, $\lambda$: 0.3) to 603.42 ($\gamma$: 0.0, $\lambda$: 0.6) and there was no clear correlation between the reward received and the parameters chosen in both runs.

The reason that the learning performance appears to be insensitive to the algorithm parameters could be a result of the implicit randomness in the game and the fact that it is a high level, simplistic design. The map used is also very small and encourages almost constant combat.
The next table shows the final kill-death difference for both runs of the experimentation. These values represent how many more kills than deaths the bot had at the end of the game. So adding 200 to these values would equal the total amount of kills carried out by the bot during the game.
\begin{table}[h]
\caption{Final Kill-Death Difference}
\begin{center}
\begin{tabular}{|c|c|c|c|c|c|}
\hline
 \bf{Run} &   & \bf{$\lambda$: 0.0} & \bf{$\lambda$: 0.3} & \bf{$\lambda$: 0.6} & \bf{$\lambda$: 0.9} \\   
\hline
1&\bf{$\gamma$: 0.0}  &162 &185 &146 &213 \\
\hline
2&\bf{$\gamma$: 0.0}  &28 &9 &245 &154 \\
\hline
\hline
1&\bf{$\gamma$: 0.3} &162 &102 &30 &122 \\
\hline
2&\bf{$\gamma$: 0.3} &154 &47 &177 &158 \\
\hline
\hline
1&\bf{$\gamma$: 0.6} &179 &100 &18 &96 \\
\hline
2&\bf{$\gamma$: 0.6} &121 &198 &173 &138 \\
\hline
\hline
1&\bf{$\gamma$: 0.9} &119 &180 &141 &108 \\
\hline
2&\bf{$\gamma$: 0.9} &128 &174 &44 &115 \\
\hline
 \end{tabular}
\end{center}
\end{table}
Again, these values are very variable and range from 18 to 213 in the first run and 9 to 245 in the second run. The number of kills is closely related to the amount of overall reward as a large reward of 1.0 is received for a kill and 0.2 for injuring an opponent. \\
\indent Five games were also played where random actions were taken 100\% of the time. The results of these games are listed in Table X below. The purpose of running these was to see what the baseline performance was like when learning was disabled and random actions were taken. The bot managed to have more kills than deaths in four out of the five games but the amount of kills and reward received was much less than when learning was enabled in the majority of the cases.
\begin{table}[h]
\caption{Random Action Games}
\begin{center}
\begin{tabular}{|c|c|c|c|}
\hline
  & \bf{Average Reward} & \bf{Total Reward} & \bf{K-D Difference}  \\   
\hline
\bf{Game 1}  &0.77&155.68&11 \\
\hline
\bf{Game 2} &0.74&149.38&8\\
\hline
\bf{Game 3}&0.84&168.57&15\\
\hline
\bf{Game 4} &0.61&122.12&-8\\
\hline
\bf{Game 5} &0.97&194.40&37\\
\hline
 \end{tabular}
\end{center}
\end{table}
The decomposition of the tasks into modes is enabling the bot to perform quite well even when selecting actions at random. For instance, when the bot is in \emph{Danger} mode, two out of the seven actions available to it involve shooting at the opponent which results in the bot being able to kill regularly. Table XI below shows the averaged values for both runs (16 games each) of average reward, total reward and kill-death difference.

\begin{table}[h]
\caption{Run 1 and Run 2 Averages}
\begin{center}
\begin{tabular}{|c|c|c|c|}
\hline
  & \bf{Average Reward} & \bf{Total Reward} & \bf{K-D Difference}  \\   
\hline
\bf{Run 1}  &1.91&382.05&128.93 \\
\hline
\bf{Run 2} &1.90&380.60&128.93\\
\hline
 \end{tabular}
\end{center}
\end{table}

We can see from this table that the performance of the bot is improved when learning is enabled in the algorithm. On average, 100 more kills are carried out and an extra 200 points of reward is received through learning. These are just preliminary runs and it will be important to carry out many more runs and continually tweak the states, actions and rewards as well as developing an improved method for evaluation in the future.

\section{Conclusion}
The paper has described an architecture for controlling NPCs in the game Unreal Tournament 2004 using multiple Sarsa($\lambda$) reinforcement learners. Early experimentation has shown that learning is occurring and good performance can be achieved. We have noted that the system is relatively insensitive to changes in the values of the RL algorithm parameters. In our future work, we will continue to investigate the reasons for this, as well as evaluating the DRE-Bot framework in more challenging environments, with larger maps and more opponents. We also plan to evaluate the performance of the bot against human opposition.



%

\vspace{-.4in}
\begin{IEEEbiography}[{\includegraphics[width=1in,height=1.2in,clip,keepaspectratio]{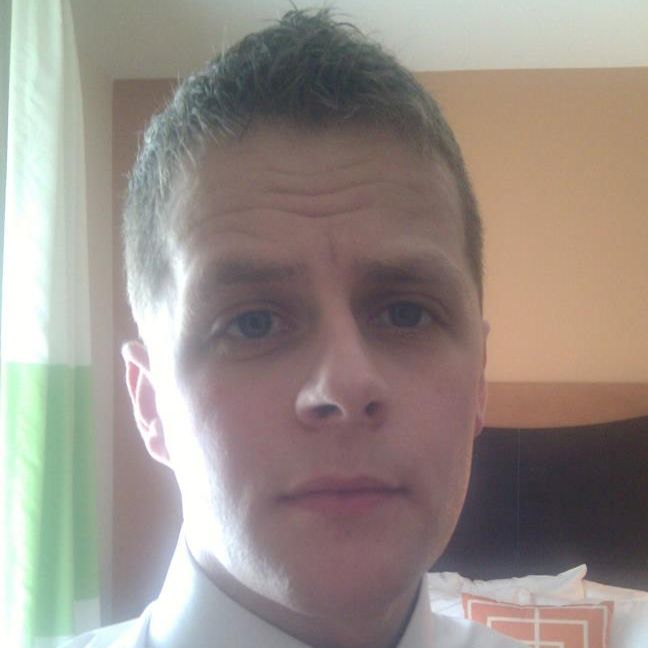}}]%
{Frank G. Glavin} 
was born in Galway, Ireland on the 7th of February 1985. He received an honours degree in Information Technology from NUI Galway in 2006. He was awarded a research MSc degree in Applied Computing and Information Technology from NUI Galway in 2010. This work involved developing a One-Sided Classification toolkit and carrying out experimentation on spectroscopy data. He is currently a PhD candidate researching the application of Artificial Intelligence techniques in modern computer games.
\end{IEEEbiography} 
\vspace{-.4in}
\begin{IEEEbiography}[{\includegraphics[width=1in,height=1.2in,clip,keepaspectratio]{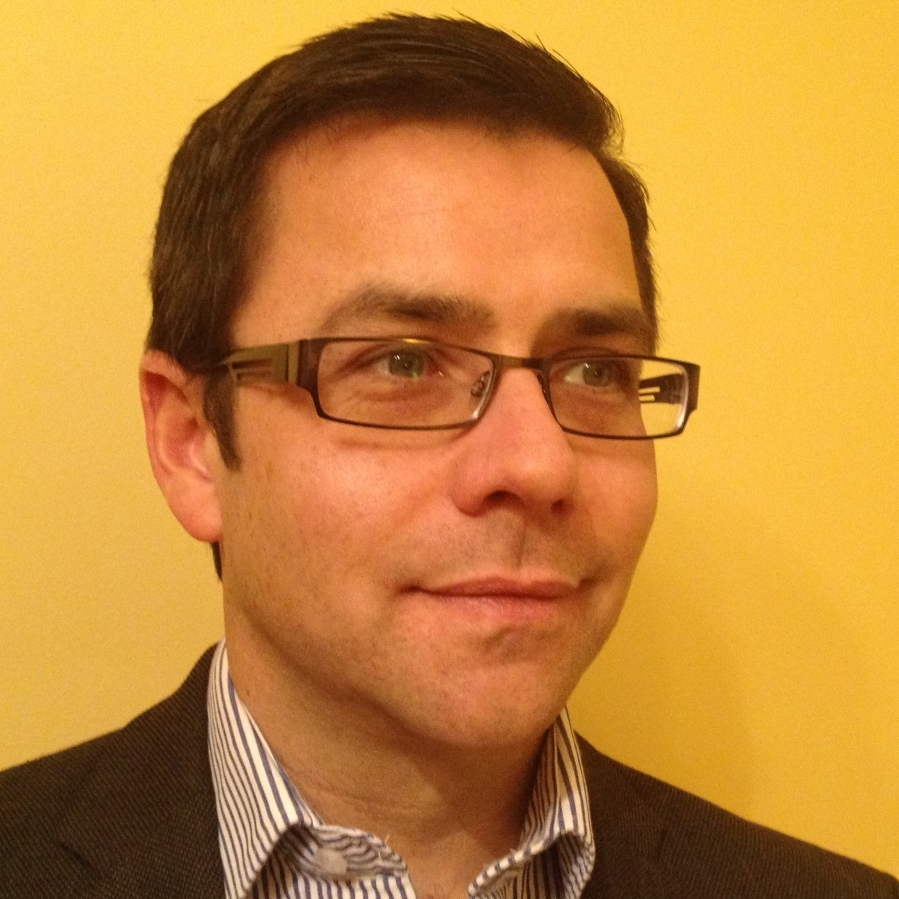}}]%
{Dr Michael G. Madden}
is the Head of the Information Technology Discipline and a Senior Lecturer in the National University of Ireland Galway, which he joined in 2000. After graduating with a B.E. from NUI Galway in 1991, he began his research career by working as a Ph.D. research assistant in Galway, then worked in professional R\&D from 1995 to 2000. He has over 80 publications, three patent filings, and co-founded a spin-out company based on his research.
\end{IEEEbiography}

\end{document}